\documentclass[letterpaper, 10 pt, conference]{ieeeconf}  

\IEEEoverridecommandlockouts                              
\usepackage{times}  
\usepackage{graphicx}
\usepackage{amsmath}
\usepackage{amssymb}
\usepackage{caption}  
\usepackage{longtable}
\usepackage{tabularx}
\usepackage{pifont}  
\usepackage{cite}
\usepackage{hyperref}

\title{\LARGE \bf
SARGes: Semantically Aligned Reliable Gesture Generation \\via Intent Chain
}

\author{Nan Gao$^{1}$, Yihua Bao$^{2}$, Dongdong Weng$^{2}$, Jiayi Zhao$^{2}$, \\Jia Li$^{1}$, Yan Zhou$^{3}$, Pengfei Wan$^{3}$, and Di Zhang$^{3}$
\thanks{$^{1}$Nan Gao and Jia Li are with the Institute of Automation, Chinese Academy of Sciences, Beijing, China.
        {\tt\small nan.gao@ia.ac.cn; jia.li@cripac.ia.ac.cn}}%
\thanks{$^{2}$Yihua Bao, Dongdong Weng and Jiayi Zhao are with the Beijing Engineering Research Center of Mixed Reality and Advanced Display, Beijing, China, and also with the Institute of Technology, Beijing, China.
        {\tt\small boye1900@outlook.com; crgj@bit.edu.cn; zjyjy@bit.edu.cn}}%
\thanks{$^{3}$Yan Zhou, Pengfei Wan, and Di Zhang are with Kuaishou Technology, Beijing, China. 
        {\tt\small zhouyan03@kuaishou.com; wanpengfei@kuaishou.com; zhangdi08@kuaishou.com}}%
}

\begin{document}

\maketitle
\thispagestyle{empty}
\pagestyle{empty}

\begin{abstract}

Co-speech gesture generation enhances human-computer interaction realism through speech-synchronized gesture synthesis. However, generating semantically meaningful gestures remains a challenging problem. We propose SARGes, a novel framework that leverages large language models (LLMs) to parse speech content and generate reliable semantic gesture labels, which subsequently guide the synthesis of meaningful co-speech gestures. First, we constructed a comprehensive co-speech gesture ethogram and developed an LLM-based intent chain reasoning mechanism that systematically parses and decomposes gesture semantics into structured inference steps following ethogram criteria, effectively guiding LLMs to generate context-aware gesture labels. Subsequently, we constructed an intent chain-annotated text-to-gesture label dataset and trained a lightweight gesture label generation model, which then guides the generation of credible and semantically coherent co-speech gestures. Experimental results demonstrate that SARGes achieves highly semantically-aligned gesture labeling (50.2\% accuracy) with efficient single-pass inference (0.4 seconds). The proposed method provides an interpretable intent reasoning pathway for semantic gesture synthesis.

\end{abstract}

\section{INTRODUCTION}

Gestures in human communication significantly enhance semantic transmission, emotional expression, and conversational flow \cite{mcneill1992hand}. Co-speech gesture generation focuses on enabling virtual characters or robots to produce synchronized, natural gestures alongside speech, advancing human-computer interaction towards greater naturalness and intelligence \cite{wei2022learning} \cite{nyatsanga2023comprehensive}. This is crucial in applications requiring enhanced realism and immersion, such as virtual agents \cite{neff2008gesture} and humanoid robots \cite{holladay2016rogue}, where natural and context-appropriate gestures greatly improve interaction quality.

Recent research on gesture generation has increasingly focused on leveraging large datasets and models. Alexanderson et al. \cite{alexanderson2023listen} showed that diffusion models are effective for synthesizing human motions, such as dancing and co-speech gestures. DiffMotion \cite{zhang2023diffmotion} integrates an autoregressive temporal encoder with a denoising diffusion model to produce high-fidelity, speech-synchronized gestures. Building on this, Diffsheg \cite{chen2024diffsheg} combines diffusion models with Transformers for real-time generation of 3D gestures and facial expressions. In multimodal applications, GestureDiffuCLIP \cite{ao2023gesturediffuclip} uses a CLIP encoder and Adaptive Instance Normalization (AdaIN) to generate a variety of gestures flexibly. MotionGPT \cite{jiang2024motiongpt} employs a pre-trained language model to interpret human motion as a foreign language, facilitating gesture generation tasks. While methods in \cite{alexanderson2023listen} and \cite{jiang2024motiongpt} can produce semantically meaningful actions like `running' and `jumping,' there is still a gap in generating semantic gestures specifically for co-speech scenarios. Moreover, despite advancements in aligning gestures with speech rhythm, diffusion models and other learning-based approaches still lack semantic understanding. Therefore, our research aims to generate meaningful gesture labels in co-speech scenarios to enhance the semantic richness of gesture generation.

Large Language Models (LLMs) are proficient at extracting semantic information and have been leveraged in gesture generation research to improve the semantic consistency of generated gestures. For instance, GesGPT \cite{gao2024gesgpt} utilizes the semantic analysis capabilities of GPT to generate meaningful gestures from text by categorizing predefined gesture intentions. However, systematic research on defining and categorizing semantic gestures for co-speech generation is limited, and much of the existing work relies on individually defined semantic gesture libraries \cite{zhang2024semantic}. Furthermore, LLMs are prone to hallucinations \cite{tonmoy2024comprehensive}, which can reduce the reliability of text-based semantic gesture generation. In ethology, ethograms are systematically organized according to specific standards to describe and understand behavior patterns \cite{stanton2015standardized}. Given that co-speech gestures are human behaviors, we adopt the concept of ethograms as a tool for gesture classification. We construct a co-speech gesture ethogram with defined guidelines and develop an LLM-based intent chain method, employing these ethogram criteria as constraints. This approach helps reduce model hallucinations and enhances the reliability of the generated gesture labels.

\begin{figure*}[htbp]
\begin{center}
\includegraphics[width=0.9\linewidth]{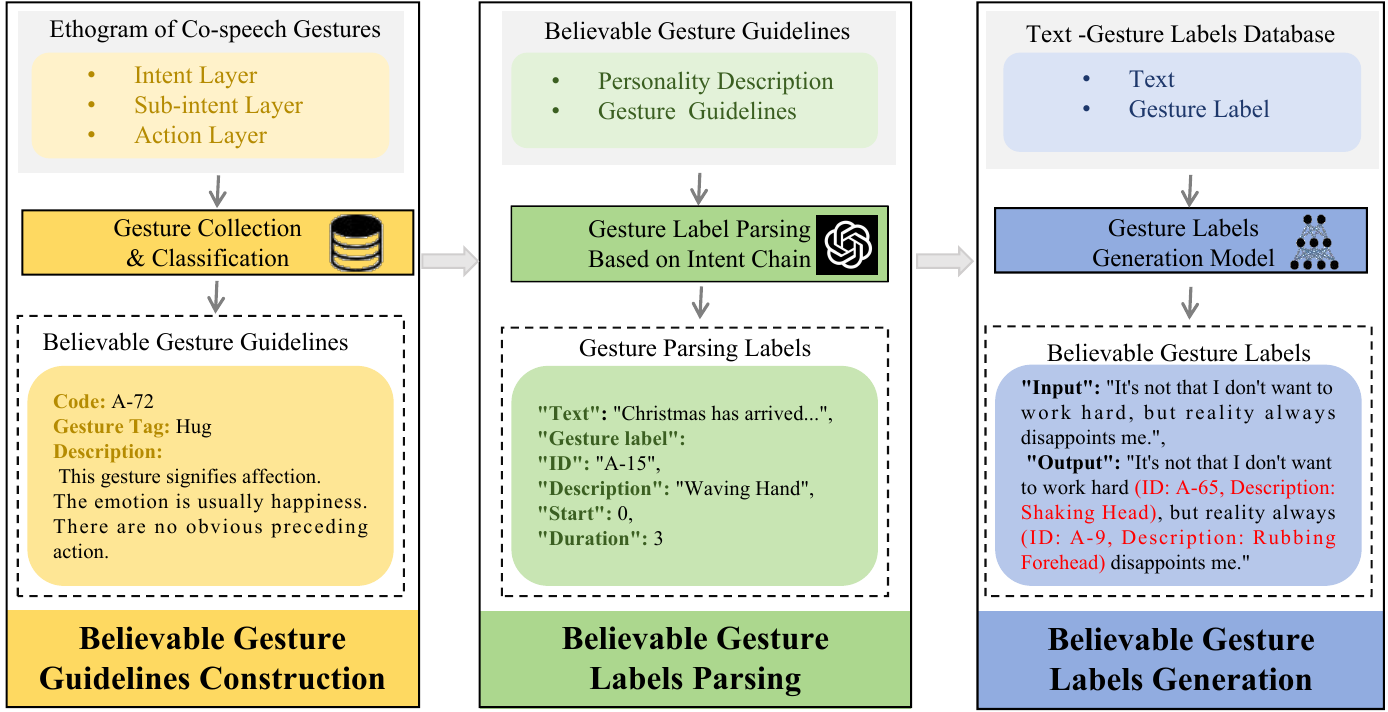}
\end{center}
\captionsetup{justification=centering, font=small} 
\caption{Pipiline for Gesture Label Generation.}
\label{fig:5}
\end{figure*}

In summary, our contributions are as follows:

\begin{itemize}
\item Inspired by research in animal behavior \cite{stanton2015standardized} and gesture cognition \cite{mcneill1992hand}, we developed a hierarchical ethogram for co-speech gestures, categorizing them based on their intentions. This ethogram serves as a systematic tool for managing semantic gestures, providing detailed descriptions of each gesture’s meaning and usage guidelines.

\item We designed an intent chain reasoning mechanism to interact with LLMs and parse gesture labels from text, utilizing techniques like chain-of-thought and self-reflection when generating gesture labels with GPT. By leveraging gesture guidelines from the ethogram as auxiliary information, we enhance the model's semantic parsing capabilities, resulting in more reliable gesture labels and reduced hallucinations.

\item We specifically constructed a dataset for gesture label generation and trained a model by fine-tuning a language model on text input. Experimental results demonstrated the effectiveness of this approach in generating appropriate gesture labels for given text, providing a reliable control signal for subsequent semantic gesture generation.

\end{itemize}

\section{RELATED WORKS}

\subsection{Semantic Gesture Generation}

Deep learning methods using multimodal inputs, like text and speech, have been employed to generate gestures by learning features from the text, providing richer semantic information \cite{kucherenko2020gesticulator}\cite{yoon2020speech}.However, these methods struggle to generate semantic gestures due to the limited examples of semantic gestures in the datasets. To further enhance the quality of semantic gesture generation, Seeg \cite{liang2022seeg} requires the predicted results to express the same semantics as the ground truth labels. The method uses semantic labels that are categorized into five classes based on the semantics and emotional characteristics of gestures.Teshima et al. \cite{teshima2022deep}, based on McNeill's work \cite{mcneill1992hand}, classify gestures into beats, imagistic, and no-gesture types, mapping input text words to corresponding gestures to generate specific ones. The Text2Gestures \cite{bhattacharya2021text2gestures} model posits that each text sentence is associated with an expected emotion, primarily represented in the VAD (Valence, Arousal, Dominance) space. Methods using motion graphs \cite{zhou2022audio}\cite{zhao2023gesture} and retrieval-based subsystems \cite{zhang2022text2video} retrieve motion segments from a predefined action library that best match the text's semantics and speech rhythm. GesGPT \cite{gao2024gesgpt} is the first to use LLMs for text intent classification, linking intent labels with semantic gestures. This method greatly improves the flexibility and accuracy of semantic gesture generation. Semantic Gesticulator \cite{zhang2024semantic} introduces a semantic-aware co-speech gesture synthesis system that utilizes a GPT-based generator along with a semantic alignment mechanism to ensure the quality and coherence of the generated semantic gestures. However, current studies often rely on custom categories or specific databases, limiting coverage. To address this, we plan to use the ethogram to create a gesture set for co-speech scenarios and explore LLM-based gesture label generation.

\subsection{LLMs in Embodied Agents}

Story-to-Motion \cite{qing2023story} employs LLMs to extract action sequences from extended texts. This method integrates action retrieval with semantic and trajectory constraints to generate character motions that are natural and controllable. In the Digital Life Project \cite{cai2024digital}, researchers proposed a framework that uses language to construct autonomous 3D characters. This approach extends action generation beyond Story-to-Motion by filtering action candidates with high-level textual semantics and refining them with kinematic features, such as joint positions, to align with the scene and character state. Another study \cite{huang2022language} translates LLM-generated action plans into executable actions by using pretrained models to match action phrases with permissible actions, ensuring semantic consistency. The Generative Agents framework \cite{park2023generative} extends LLMs to store comprehensive experience records of agents, enabling them to form higher-level reflections based on these memories and dynamically retrieve them for behavior planning. This approach allows agents to exhibit realistic social behaviors in simulated environments, enhancing the authenticity of interactions. Collectively, these works demonstrate that LLMs can effectively parse human behavior-related information from text and convert it into executable actions,  thereby serving as a viable approach for achieving semantically gesture generation.

\section{PROPOSED METHOD}

This study proposes a method using LLMs to generate gesture labels from text. Given a text \( x \), the method generates the corresponding gesture label \( y \), where \( y \) belongs to a predefined gesture ethogram \( G \). This relationship is modeled by \( y = M_{\Phi}(x) \), with \( y \in G \).

\subsection{Pipeline} Firstly, as shown in Fig.1, we established a gesture ethogram for co-speech scenarios, which includes a gesture classification hierarchy and usage guidelines. These guidelines enhance the reliability of gesture labels by providing clear guidance for GPT text parsing. We designed an intent chain reasoning method based on LLMs for behavior intention inference, employing chain-of-thought \cite{wei2022chain} and self-reflective \cite{shinn2023reflexion} strategies to generate gesture label parsing results aligned with the text. Subsequently, based on the parsed labels, we constructed a dataset to train a gesture label generation model. These labels can then be used in conjunction with motion matching or motion generation techniques to enhance semantic gesture generation in applications such as virtual digital humans and social robots. 

\subsection{Believable Gesture Guidelines Construction}
This section utilizes the ethogram construction methodology to categorize and collate gestures commonly observed in co-speech scenarios. We then provide comprehensive descriptions of each categorized gesture, along with specific usage guidelines. 

\subsubsection{Ethogram of Co-speech Gestures}
In ethological research, ethograms are commonly used to identify patterns and regularities in animal behavior \cite{stanton2015standardized}\cite{lehner1987design}. We plan to construct an ethogram for co-speech gestures, systematically classifying and organizing gestures that occur in co-speech scenarios. Following the hierarchical construction strategy used in ethological research  \cite{lehner1987design}, we organize the ethogram of co-speech gestures into three layers: the Intent Layer, the Sub-intent Layer, and the Action Layer. This layered structure effectively transitions from behavioral motivations to specific gestures.

\textbf{Intent Layer:}
This layer provides a high-level abstraction of the motivations behind co-speech gestures. Based on McNeill's classification of gesture functions \cite{mcneill1992hand}, we categorize gesture motivations into four types: \textit{Information Display}, \textit{Concrete Reinforcement}, \textit{Tone Reinforcement}, and \textit{Comfort Behaviors}.

\textit{Information Display} gestures, similar to what McNeill describes as emblematic gestures, convey meanings on their own, separate from speech, like showing emotions or depicting certain looks. \textit{Concrete Reinforcement} includes iconic, metaphoric, and pointing gestures, which help verbal communication by showing attributes like direction, shape, or position. Beat gestures fit under \textit{Tone Reinforcement}, as they express emotions through emphasis, rhythm, or a questioning tone. \textit{Comfort Behaviors} are gestures that help alleviate emotions such as tension, anxiety, or sadness. These gestures are also common in human-computer interaction; therefore, we have included this category in the intent layer.

\textbf{Sub-intent Layer:}
This layer refines the Intent Layer by breaking down broader intents into more specific sub-intents. For instance, within the \textit{Comfort Behaviors} category, sub-intents can include alleviating tension, fear, anxiety, excitement, boredom, sadness, embarrassment, and anger, among others. 

\textbf{Action Layer:}
This layer focuses on specific gestures that directly reflect the intents of the corresponding layers. For example, gestures like \textit{spreading arms wide}, \textit{waving hands}, and \textit{two-handed applause}. The complete gesture spectrum can be found at \href{https://github.com/gesture-label/ethogram}{https://github.com/gesture-label/ethogram}.


By collecting and classifying a wide range of gestures from communication scenarios according to the defined ethogram, we can systematically organize gestures in co-speech contexts. New gestures can be categorized within the ethogram based on its definitions, ensuring a clear rationale for classification. This approach allows for continuous refinement as more data and new gestures are added. Additionally, we assign an index to gestures based on their distinct motivations, enabling precise management and retrieval of gesture data, as illustrated in Table 1.

\vspace{-0.5em}

\begin{table}[h]
\captionsetup{font=small, justification=centering} 
\caption{Ethogram of Co-speech Gestures} 
\label{table:intention_layers}
\centering
\renewcommand{\arraystretch}{1.5} 
\scalebox{0.55}{%
\begin{tabular}{c|c|c} 
\hline
\multicolumn{1}{c|}{\Large \textbf{Intent Layer} \rule[-1ex]{0pt}{3ex}} & \Large \textbf{Sub-intent Layer} & \Large \textbf{Action Layer} \\ \hline
& \textit{\large\addtolength{\spaceskip}{1pt} Display Appearance} & \textit{\large\addtolength{\spaceskip}{1pt} A-1: Stretch Shoulders} \\ 
\textbf{\large\addtolength{\spaceskip}{1pt} Information Display} & \textit{\large\addtolength{\spaceskip}{1pt} Display Special Meaning} & \textit{\large\addtolength{\spaceskip}{1pt} A-2: Thumbs Down } \\ 
& \textit{\large\addtolength{\spaceskip}{1pt} ...} & \textit{\large\addtolength{\spaceskip}{1pt} ...} \\ \hline
& \textit{\large\addtolength{\spaceskip}{1pt} Concrete Direction} & \textit{\large\addtolength{\spaceskip}{1pt} B-1: Point Finger in Target Direction} \\ 
\textbf{\large\addtolength{\spaceskip}{1pt} Concrete Reinforcement} & \textit{\large\addtolength{\spaceskip}{1pt} Concrete Shape} & \textit{\large\addtolength{\spaceskip}{1pt} B-2: Form Hands into a Circle} \\ 
& \textit{\large\addtolength{\spaceskip}{1pt} ...} & \textit{\large\addtolength{\spaceskip}{1pt} ...} \\ \hline
& \textit{\large\addtolength{\spaceskip}{1pt} Questioning Tone} & \textit{\large\addtolength{\spaceskip}{1pt} C-1: Wave Palm Upwards} \\ 
\textbf{\large\addtolength{\spaceskip}{1pt}  Tone Reinforcement} & \textit{\large\addtolength{\spaceskip}{1pt} Emphasizing Tone} & \textit{\large\addtolength{\spaceskip}{1pt} C-2:Shake Interlocked Fists} \\ 
& \textit{\large\addtolength{\spaceskip}{1pt} ...} & \textit{\large\addtolength{\spaceskip}{1pt} ...} \\ \hline

& \textit{\large\addtolength{\spaceskip}{1pt} Soothe Nervousness} & \textit{\large\addtolength{\spaceskip}{1pt} D-1: Rub or Pinch Fingers} \\ 
\textbf{\large\addtolength{\spaceskip}{1pt} Comfort Behaviors} & \textit{\large\addtolength{\spaceskip}{1pt} Soothe Fear} & \textit{\large\addtolength{\spaceskip}{1pt} D-2:Cover Eyes with Hands} \\ 
& \textit{\large\addtolength{\spaceskip}{1pt} ...} & \textit{\large\addtolength{\spaceskip}{1pt} ...} \\ \hline
\end{tabular}
}
\end{table}

\vspace{-0.5em} 

\subsubsection{Believable Gesture Guidelines }

Based on the co-speech gesture ethogram, we collected over 200 co-speech gestures. We will establish guidelines by analyzing their usage and identifying keywords frequently associated with each gesture. These guidelines will serve as auxiliary information in subsequent LLM-based gesture label parsing, enhancing the reliability of generating gesture labels corresponding to text.

We initially employed prompt engineering with ChatGPT to generate usage descriptions for all gestures within the ethogram, focusing on contextual and emotional correlations. To enhance output consistency, example texts were provided as inputs. Following the generation of descriptions, we conducted manual reviews and corrections to ensure accuracy and uniformity. This process culminated in the development of the Believable Gesture Guidelines, which provide a robust theoretical foundation for gesture label generation. Fig. 2 offers a partial illustration of gesture guideline. 



\begin{figure}[htbp]
\begin{center}
\includegraphics[width=0.9\linewidth]{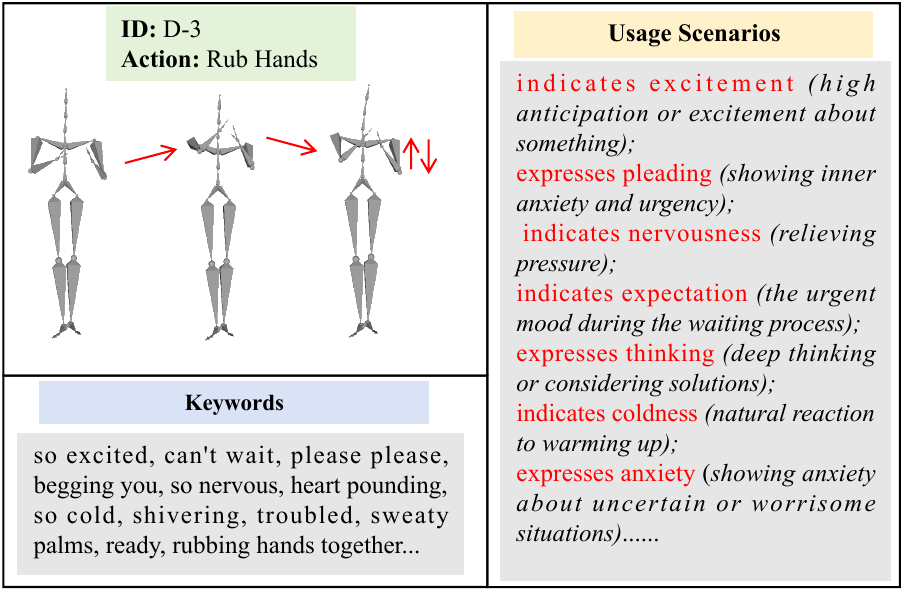} 
\end{center}
\captionsetup{font=small, justification=centering} 
\caption{Guidelines Illustration for the `Rub Hands' Gesture}
\label{fig:5}
\end{figure}

\subsection{Believable Gesture Labels Parsing }
By applying gesture guidelines as constraints, we ensure that the labels parsed from the intent chain based on LLMs are appropriate and believable for the given text.

\subsubsection{Chain-of-Thought Gesture Prompting}
Chain-of-thought (CoT) prompting improves model reasoning by encouraging intermediate steps in problem-solving \cite{wei2022chain}. We designed prompts using the CoT approach to interact with ChatGPT. Our aim is for the model to process text segments step-by-step, identify relevant keywords, and select appropriate gestures from the ethogram based on predefined guidelines.

First, we establish the virtual agent's character profile as a foundation for LLM interactions. Then, the LLM clarifies the conversation theme and identifies the speaker's primary intent. Based on this intent, the model analyzes keywords from the text and, considering the character's personality, selects appropriate gestures from the defined gesture guidelines. These gestures are then inserted before the identified keywords. This approach allows the language model to select the most suitable gestures, guided by character traits and context. By clearly defining each step and decision criterion, it helps reduce the occurrence of model hallucinations and enhances the credibility of text parsing.

\subsubsection{Gesture Prompting with Self-reflection}
Self-reflection can significantly enhance decision-making capabilities in agent interaction tasks \cite{shinn2023reflexion}. In our research, we integrate this mechanism into prompt engineering. Utilizing the CoT prompting strategy, we introduce multiple rounds of gesture selection and reflection, allowing LLMs to re-evaluate and refine their results based on established rules. The reflection rules encompass semantic relevance and action relevance.

The semantic relevance evaluation includes: (i) Context Matching: Checking if body movements align with the speaker's identity, theme, and setting, such as using open gestures in positive contexts or composed postures in serious discussions. (ii) Keyword Matching: Assessing if gestures semantically match the keywords. (iii) Consistency of Emotional Expression: Evaluating if gestures align with the emotional tone, like using faster movements for exciting or joyful topics.

The action relevance evaluation includes: (i) Positional Consistency: Evaluating whether the placement of gestures aligns with the position of the keywords. (ii) Moderation: Evaluating the frequency of gestures to ensure they are not too frequent, with no more than two gestures in a single sentence.

\subsubsection{Gesture Parsing Labels}

Each gesture label includes: a unique Gesture ID, such as `A-15' for the 15th gesture in \textit{Information Display}; the gesture's name; its start position in the text, measured in characters (e.g., `0' for the beginning); the duration, also in characters (e.g., `5' for five characters).

\subsection{Believable Gesture Labels Generation}
We employed GPT to parse gesture labels and build a text-gesture labels dataset. This dataset facilitated the training of a model focused on reliable, cost-efficient gesture label generation.

\subsubsection{Text-Gesture Labels Database}
We collected various speech videos from the internet and utilized speech-to-text technology to extract the corresponding text, forming a corpus in which each sentence is treated as a distinct unit. Using the gesture label parsing method previously discussed, we annotated this corpus with gesture labels. The labels are formatted as follows:

\small
\textit{{Input: "Hello, it's great to have you here today. You are truly amazing!" \\
Output: "Hello, it's great (id: A-97, description: spreading arms wide) to have you here today. You are truly (id: A-6, description: clapping) amazing!"}}
\normalsize

Through this approach, we constructed the training dataset for the gesture label generation model.

\begin{figure*}[htbp]
\begin{center}
\includegraphics[width=0.9\linewidth]{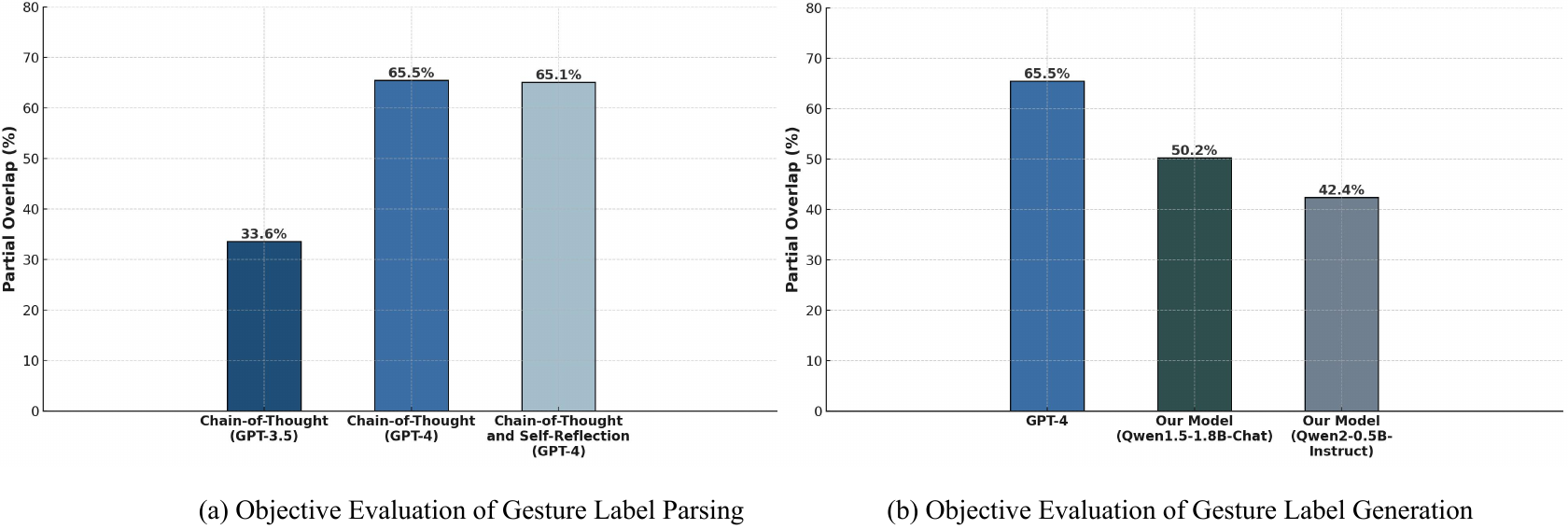} 
\end{center}
\captionsetup{justification=centering, font=small} 
\caption{Objective Evaluation Results}
\label{fig:5}
\end{figure*}

\subsubsection{Gesture Labels Generation Model}
LLMs trained on vast datasets have demonstrated exceptional performance in various natural language processing tasks, including translation \cite{xu2024contrastive}. We conceptualize the training of the gesture label generation model as a text translation task, where the input consists of the current dialogue text, and the output is the text annotated with gesture labels. We fine-tune open-source LLMs to exploit their robust semantic processing capabilities, thereby enhancing their adaptability to the gesture label generation task.

For the gesture label generation task, the training dataset is defined as: \( Z = \{(x_i, y_i)\}_{i=1,..,N} \), where \(x_i\) is the natural query input, and \(y_i\) is the text with gesture labels. We fine-tune a language model Qwen based on the transformer architecture \cite{xu2024contrastive}, represented as \(M_{\Phi_0}(y|x)\), with parameters \(\Phi_0\). LoRA (Low-Rank Adaptation) is a method that achieves model compression and acceleration through low-rank decomposition \cite{bai2023qwen}. This approach uses low-rank matrices to replace or adjust the weight matrices of the original model, achieving efficient parameterization and significantly reducing resource consumption in training and inference. We apply LoRA technique \ to the q\_proj and v\_proj modules of the Qwen model by inserting low-rank matrices to adjust their outputs. The goal is to maximize the conditional log-likelihood of the language model, enabling it to generate gesture labels. The optimization problem can be expressed as (1):

\begin{equation}
\max_{\Theta} \sum_{(x, y) \in Z} \sum_{i=1}^{|y|} \log \left(M_{\Phi_0 + \Delta\Phi(\Theta)}(y_i \mid x, y_{<i})\right)
\end{equation}

where \(M_{\Phi_0 + \Delta\Phi(\Theta)}(y_t \mid x, y_{<i})\) represents the conditional probability computed with the current parameters \(\Phi_0 + \Delta\Phi(\Theta)\). To improve computational and memory efficiency, \(\Delta\Phi(\Theta)\) can utilize the low-rank representation proposed by the LoRA method, significantly reducing the number of parameters and computational load during training.

\section{Experiments}
\subsection{Gesture Label Parsing}
\textbf{Test Dataset:}
We designed a specific text test dataset based on gestures from the created ethogram. Specifically, we selected 77 representative gestures from the ethogram and constructed relevant text examples according to the semantic descriptions of each gesture. For example, the gesture `rubbing hands,' described as `expressing excitement or pleading,' is paired with the example sentence, \textit{`I can hardly wait, this news has me so excited!'} which is semantically aligned with the action of `rubbing hands.' Similarly, we designed three highly related example sentences for each gesture, resulting in a total of 231 sentences that make up the test dataset. These sentences are paired with their corresponding gesture labels and are used to evaluate the performance of the prompt-based method in parsing the alignment between text and gesture labels.

\textbf{Evaluation Metric:}
We designed the \textit{Partial Overlap} metric to measure the accuracy of the gesture labels $\hat{y}$ in the parsing results relative to the true labels $y$, as defined by Equation (1). Given that we include 77 gesture labels, with a sentence potentially corresponding to multiple labels, we addressed this complexity by clustering the labels into five emotion-based categories: `joy,' `anger,' `sorrow,' `fear,' and `special meaning.' The `special meaning' category includes gestures with unique meanings, such as `bunny ears.' In accuracy evaluation, we map both the gesture label parsing results and the test data labels to these five categories and then calculate the Partial Overlap on the mapped results using (2).

\begin{equation}
\text{Partial Overlap} = \frac{\sum_{i=1}^{N} |\hat{y}_i \cap y_i|}{\sum_{i=1}^{N} |y_i|}
\end{equation}

We employed CoT and self-reflection prompt engineering methods for text-based gesture label parsing and compared the text parsing performance between GPT-3.5 and GPT-4 models. In the self-reflection prompting method, the results were iterated three times based on the reflection criteria mentioned above. The results are shown in Fig. 3.

Based on the experimental results, it is clear that the accuracy of gesture label parsing using GPT-3.5 is substantially lower compared to GPT-4, highlighting the necessity of a more advanced model for enhanced semantic comprehension in this task. Specifically, using GPT-4 for annotation yielded an accuracy of 65.5\%. However, incorporating self-reflection resulted in a marginal decrease in accuracy compared to the CoT prompt method.

In tasks with objective evaluation criteria, such as programming (e.g., successful code compilation) \cite{nam2024using}, self-reflection can effectively enhance the model's decision-making process. However, for gesture generation, an ill-posed problem lacking clear evaluation standards, the reflection mechanism can introduce additional uncertainty, potentially leading to increased variability in the results and limiting improvements in decision-making accuracy.


\begin{figure*}[htbp]
\begin{center}
\includegraphics[width=0.9\linewidth]{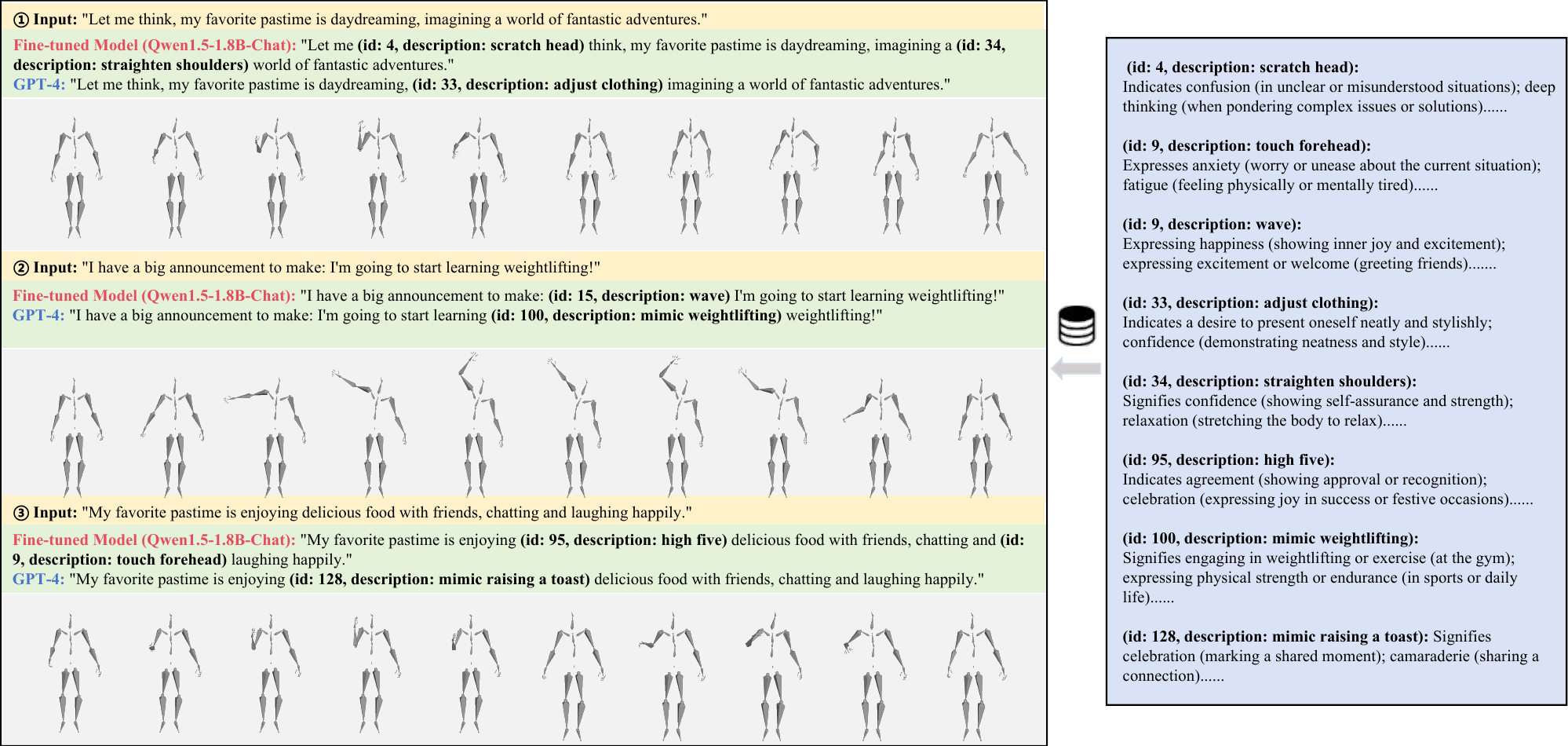} 
\end{center}
\captionsetup{justification=centering, font=small} 
\caption{Gesture Label Generation Visualization. It is important to note that we did not further distinguish different categories (such as A, B, C, and D) within the gesture IDs in order to avoid increasing the difficulty of model training.}
\label{fig:5}
\end{figure*}

\subsection{Gesture Label Generation}
\subsubsection{Objective Evaluation}

We employed the CoT prompting technique using GPT-4 to parse text for the construction of a training dataset, resulting in a training dataset comprising 3,242 entries. Each entry includes the original text with embedded gesture labels formatted as \textit{(ID: Action Name)}, facilitating direct integration of gesture annotations within the text.

\textbf{Experimental Parameters:}
We utilized Qwen  \cite{bai2023qwen} for fine-tuning. The LoRA technique was applied to the \( q\_proj \) and \( v\_proj \) layers, with a batch size of 6, a learning rate of \( 5 \times 10^{-5} \), over 100 epochs. Training was performed on a 4090 server, with a total duration of approximately 13 hours.For evaluation, we used the same test dataset and metrics as previously, involving 231 test texts and employing Partial Overlap as the primary evaluation metric. The results of the experiment are presented in Fig. 3.

Compared to the performance of GPT-4 in generating gesture labels (with an overlap rate of 65.5\%), our model also demonstrated the capability to produce semantically relevant action labels. Specifically, the fine-tuned models based on Qwen1.5-1.8B-Chat and Qwen2-0.5B-Instruct achieved Partial Overlap rates of 50.2\% and 42.4\%, respectively, on the test set. Our gesture label generation model, despite utilizing a smaller dataset, still achieves comparable accuracy in generating gesture labels that align with the text semantics, slightly lower to GPT-4.

Moreover, the comparison of response times revealed a significant performance advantage of the our model. GPT-4's average response time is about 3 seconds, processing approximately 8,000 tokens to generate a single set of action labels, with a cost of \$0.26. In contrast, the response time of our models was significantly reduced to about 0.4 seconds, greatly enhancing generation efficiency. Although the Partial Overlap rate of the fine-tuned models is slightly lower, their faster response time and significantly reduced cost offer advantages in practical applications, particularly in scenarios demanding high response speed and lower resource consumption.

\subsubsection{Discussion}
We present specific examples of gesture label generation, as illustrated in Fig. 4. The left column shows the generated labels, each accompanied by the corresponding gestures produced by our model using a database retrieval approach. The parsing results from GPT-4 are used as a benchmark for our comparative analysis. In Fig. 4, the left side \ding{172}. the input \emph{`Let me think, my favorite pastime is daydreaming, imagining a world of fantastic adventures.'} prompted our model to insert a \emph{`scratch head'} gesture at \emph{`think'}, reflecting thought and contemplation. Both our model and GPT-4 generated gestures indicating comfort in \emph{`imagining a world...'}, demonstrating semantic consistency. 

In Fig. 4, the left side \ding{173} shows the input \emph{`I have a big announcement, I am going to learn weightlifting,'} for which GPT-4 accurately generated the \emph{`id: 100: mimic weightlifting'} gesture. In contrast, our model inserted a \emph{`wave'} gesture that, although not semantically aligned, effectively conveyed a joyful emotion. This demonstrates that GPT-4 excels in deep semantic understanding, while our model can achieve similar semantic gesture label recognition at a lower cost.

In hard cases, such as in Fig. 4, the left side \ding{174}, where the input conveyed a joyful mood during a meal with friends, the model generated a \emph{`id: 9, description: touch forehead'} gesture, which is typically associated with confusion or anxiety, resulting in a mismatch with the intended context.

Overall, our gesture label generation method demonstrates satisfactory performance, particularly with advantages in cost and efficiency. However, compared to GPT-4, the fine-tuned model shows limitations in semantic depth and detail sensitivity, occasionally resulting in less accurate gesture choices. These issues could be improved by further increasing the diversity of the training data.

\section{CONCLUSION}

In this paper, we propose a novel framework for generating gesture labels from text, laying the foundation for semantic gesture generation. We developed a gesture ethogram tailored for co-speech scenarios, systematically classifying gesture patterns. We designed an intent chain behavior intention parsing method based on LLMs by integrating reliable gesture guidelines and advanced prompting techniques, allowing us to parse text and generate corresponding gesture labels effectively. Using these labels, we constructed a dataset and fine-tuned a language model, enabling the effective generation of semantically meaningful gesture labels. Although our model's accuracy is slightly lower than that of GPT-4, it demonstrates significant advantages in response speed and cost efficiency, making it suitable for real-time applications in virtual agents and social robots.

Our approach effectively generates reliable, fast, and low-cost gesture labels, with potential for further performance improvements. Future work will focus on expanding training data diversity, enhancing gesture label accuracy, and applying these labels in subsequent semantic gesture generation tasks.


\section*{ACKNOWLEDGMENT}
This work was supported by the National Key R\&D Program of China under Grant 2022YFF0902202.


\end{document}